%% file: main.tex
\documentclass[10pt,twocolumn,letterpaper]{article}
\usepackage[pagenumbers]{cvpr}
\input{preamble}
\definecolor{cvprblue}{rgb}{0.21,0.49,0.74}
\usepackage[pagebackref,breaklinks,colorlinks,allcolors=cvprblue]{hyperref}
\def\confName{CVPR}
\def\confYear{2027}
\title{Semantic Role--Structured World Model for Physical Interaction Planning}
\author{%
Juntao Cheng\textsuperscript{1}\quad
Yijun Shen\textsuperscript{2}\quad
Qunchao Jin\textsuperscript{3}\quad
Xiansheng Chen\textsuperscript{4}\quad
Wei Liu\textsuperscript{4}\\
Rui Qian\textsuperscript{5}\quad
Jingyi Peng\textsuperscript{6}\quad
Yunsong Zhou\textsuperscript{7}\quad
Zhiwei Yu\textsuperscript{4}\\[0.4em]
{\normalsize \textsuperscript{1}Shanghai Jiao Tong University\quad
\textsuperscript{2}East China Normal University}\\
{\normalsize \textsuperscript{3}University of Adelaide\quad
\textsuperscript{4}Beijing Academy of Artificial Intelligence}\\
{\normalsize \textsuperscript{5}Fudan University\quad
\textsuperscript{6}Singapore Management University}\\
{\normalsize \textsuperscript{7}Shanghai Artificial Intelligence Laboratory}%
}

\begin{document}
\maketitle
\begin{abstract}
World models for robotic manipulation must capture how actions transform a scene in ways that are meaningful for task execution.
Existing object-centric approaches preserve the transient identity of entities, while leaving the evolving task structure that gives those dynamics meaning largely implicit.
We introduce \method{}, a world model that represents scene evolution through semantic roles: a task-conditioned structure explicitly encoding the functional status of scene entities, their task-relevant relations, and the current phase of execution.
Conditioned on language, visual observations, and proprioception, \method{} uses this structured state to govern action generation and action-conditioned dynamics, predicting interaction events, relation satisfaction, and phase progression.
At inference time, these predictions guide phase-aware action selection and local repair of invalid rollouts.
We instantiate \method{} on both lightweight \backbone{} and pretrained \cosmos{} backbones, improving closed-loop success from 45.5\% to 83.0\% on LIBERO, 42.9\% to 55.2\% on RoboCasa, and 55.0\% to 81.0\% on real-robot Galbot transfer.
Across backbones and domains, explicitly modeling task-semantic structure yields substantially stronger world-model-based manipulation.
\end{abstract}
\input{sections/01_introduction}

\input{sections/02_related}
\input{sections/03_method}
\raggedbottom
\input{sections/04_experiments}
\flushcolsend
\input{sections/05_conclusion}
\clearpage
\raggedcolsend

\input{sections/06_statements}
{
    \small
    \bibliographystyle{ieeenat_fullname_unsorted}
    \bibliography{main}
}

\input{supplementary}

\end{document}

%% file: preamble.tex
\usepackage[T1]{fontenc}
\usepackage{bm,array,tabularx}
\usepackage{placeins}
\usepackage[nospread,checkfloat]{flushend}
\raggedend
\ifdefined\pdfminorversion
\fi
\newcommand{\method}{\textup{\textsc{SemRole}}}
\newcommand{\backbone}{\textup{\textsc{LeWM}}}
\newcommand{\cosmos}{\textup{\textsc{Cosmos}}}

\newcommand{\tabstyle}{\small\setlength{\tabcolsep}{4pt}\renewcommand{\arraystretch}{1.10}}

%% file: sections/01_introduction.tex
\section{Introduction}
\label{sec:introduction}

World models enable agents to anticipate the consequences of actions and evaluate possible futures before acting \citep{hafner2025mastering,zhou2024dino,maes2026leworldmodel}. 
For manipulation, these consequences include changes in both the observable scene and its task-semantic structure. 
Beyond changes in appearance, geometry, or motion, actions alter how entities participate in an instructed interaction: their \emph{functional status}, the \emph{task-relevant relations} they satisfy, and the \emph{execution phase} in which those relations matter. 
We call this coupled, instruction-dependent organization \emph{semantic-role structure}. 
A world model for physical interaction planning should connect predicted physical changes to this structure, making explicit whether they establish, preserve, or violate the conditions needed for task progress.

\begin{figure*}[t]
\centering
\includegraphics[width=\linewidth]{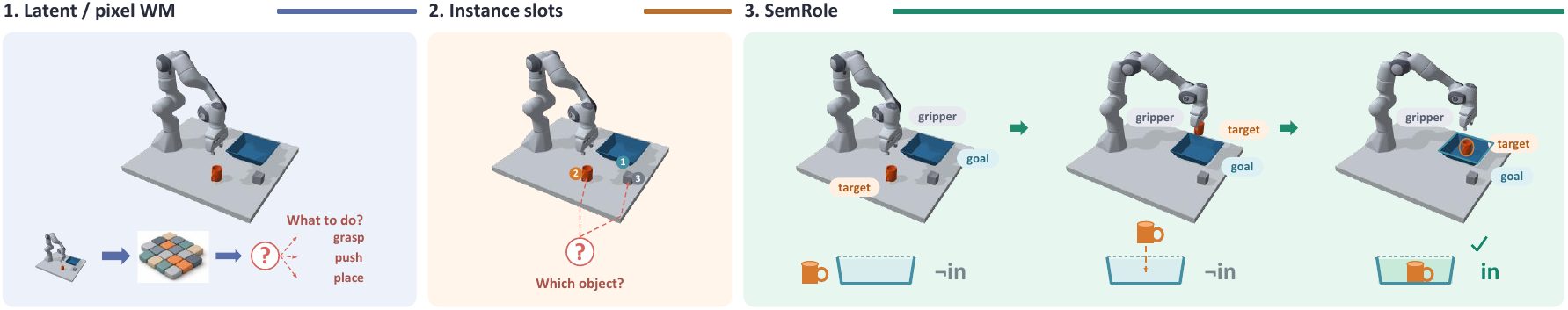}
\caption{\textbf{Comparison of world-model representations for task-conditioned manipulation.}
(a) Latent/pixel world models capture visual dynamics without explicitly representing task-relevant roles.
(b) Instance-centric representations distinguish scene entities but do not explicitly specify their functional roles in the instructed task.
(c) \method{} jointly represents task participants (gripper, target, and goal), execution phases (approach, grasp, and place), and required relational transitions, enabling structured prediction of task progress.}
\label{fig:intro-comparison}
\end{figure*}

Object-centric world models provide a useful starting point: they represent scene entities and their interactions, supporting structured prediction and control \citep{kipf2020contrastive,ferraro2025focus,spieler2026slot}. However, an entity-centered state does not by itself specify which interactions an instruction requires at each stage. Figure~\ref{fig:intro-comparison} makes this distinction concrete. Opening a gripper ends a grasp in both cases, but during transport it violates a required relation, whereas after support is established at the goal it completes a valid release. Predicting this physical effect must therefore be coupled with determining whether it establishes or preserves the conditions needed for task progress.

Task objectives can turn such predictions into decisions, as demonstrated by reward-based imagination \citep{hafner2025mastering} and visual-goal planning \citep{zhou2024dino}. However, a return or goal-distance score alone does not identify \emph{which interaction condition fails} or \emph{where an invalid continuation begins}. Without this information, rejecting a candidate does not specify how to retain its useful prefix and correct the remainder. Selection also cannot recover a suitable action absent from the proposal set. This motivates a task representation that shapes proposed actions and exposes intermediate interaction conditions for their evaluation and repair.

We introduce \method{}, a \emph{semantic role--structured world model for physical interaction planning} (Figure~\ref{fig:intro-comparison}(c)). \method{} makes semantic-role structure the organizing principle of a shared predictive \emph{role state}. Concretely, we bind visual entities to their task functions, couple them with relational and phase context, and learn action-conditioned transitions in this joint space. These transitions expose whether actions establish, preserve, or violate the interaction conditions required for progress. The same state also conditions action generation, so task structure shapes candidates before their predicted consequences are evaluated.

To ground task participation in observations, an instruction-conditioned binder combines visual entities with language and proprioception, while an interaction-history encoder supplies phase context. To improve the proposals available to the selector, we condition a flow-matching action generator on the same role state. To test whether candidates respect changing requirements, a semantic rollout module predicts intermediate relation and phase transitions for phase-aware selection. When these predictions locate the first violation, a prefix-conditioned repair module preserves the preceding action sequence and regenerates only its suffix. New observations update the role state after each executed prefix, grounding the next decision in observed execution.

We instantiate \method{} with a compact \backbone{} backbone on LIBERO and pretrained \cosmos{} on RoboCasa and Galbot. On LIBERO, role conditioning improves single-candidate success from 45.5\% to 62.8\%, with selection and repair increasing it to 83.0\%. Within the \cosmos{} setting, success rises from 42.9\% to 55.2\% across 17 RoboCasa tasks and from 55.0\% to 81.0\% on Galbot.

Our contributions are threefold: \textbf{(i)} we introduce \method{}, a semantic role--structured world model that makes task participation a shared basis for action generation and physical interaction planning; \textbf{(ii)} we develop instruction-conditioned participant binding, phase-aware semantic rollout evaluation, and prefix-conditioned suffix repair to ground task functions, distinguish progress from violations, and correct invalid continuations; \textbf{(iii)} we evaluate the model on LIBERO, RoboCasa, and Galbot across two backbones, with component studies that distinguish gains in generation, selection, and repair.

%% file: sections/02_related.tex
\section{Related Work}
\label{sec:related}

\paragraph{World models for planning.}
World models use imagined trajectories or visual prediction to support action decisions \citep{hafner2025mastering,zhou2024dino,maes2026leworldmodel,cheng2026causalwam}. \cosmos{} provides pretrained foundation-scale world models \citep{nvidia2025cosmos}. Our state instead exposes task entities, relations to establish or preserve, and execution phase.

\paragraph{Object-centric interaction models.}
C-SWM and Slot Attention learn relational or exchangeable object representations \citep{kipf2020contrastive,locatello2020object}; FOCUS, VIOLA, and Slot-MPC use object structure for exploration, control, and planning \citep{ferraro2025focus,zhu2022viola,spieler2026slot}. OA-WAM connects addressable objects to an action decoder \citep{liu2026oa}; WorldDP combines object-centric subgoal planning with diffusion policies \citep{goswami2026unifying}; MaskWAM unifies mask prompting and prediction \citep{yu2026maskwam}. Object-centric Gaussian world models predict action-conditioned dynamics \citep{kreber2026learning}; \method{} shares entity functions, relations, and phase across generation and prediction, with Grounding DINO/SAM proposals as optional visual evidence \citep{liu2024grounding,kirillov2023segment,ravi2025sam}.

\paragraph{Task structure in manipulation.}
Task and motion planning links discrete structure to geometric constraints \citep{garrett2021integrated}. SWITCH benchmarks state tracking and outcome verification in long-horizon interface interactions \citep{cheng-etal-2026-switch}. \method{} links predicted relations and phase to candidate generation and selection.

\paragraph{Generative actions and closed-loop decisions.}
Diffusion Policy, Diffuser, and flow matching model multimodal sequences through receding-horizon denoising, inpainting, or continuous transport \citep{chi2025diffusion,janner2022planning,lipman2023flow}. \method{} builds on conditional generation; its contribution is the shared role state and the use of predicted semantic transitions to select candidates and localize suffix repair. Our experiments distinguish action-prior quality, selection accuracy, and closed-loop success.

%% file: sections/03_method.tex
\begin{figure*}[t]
\centering
\includegraphics[width=\linewidth,trim=0 3bp 0 3bp,clip]{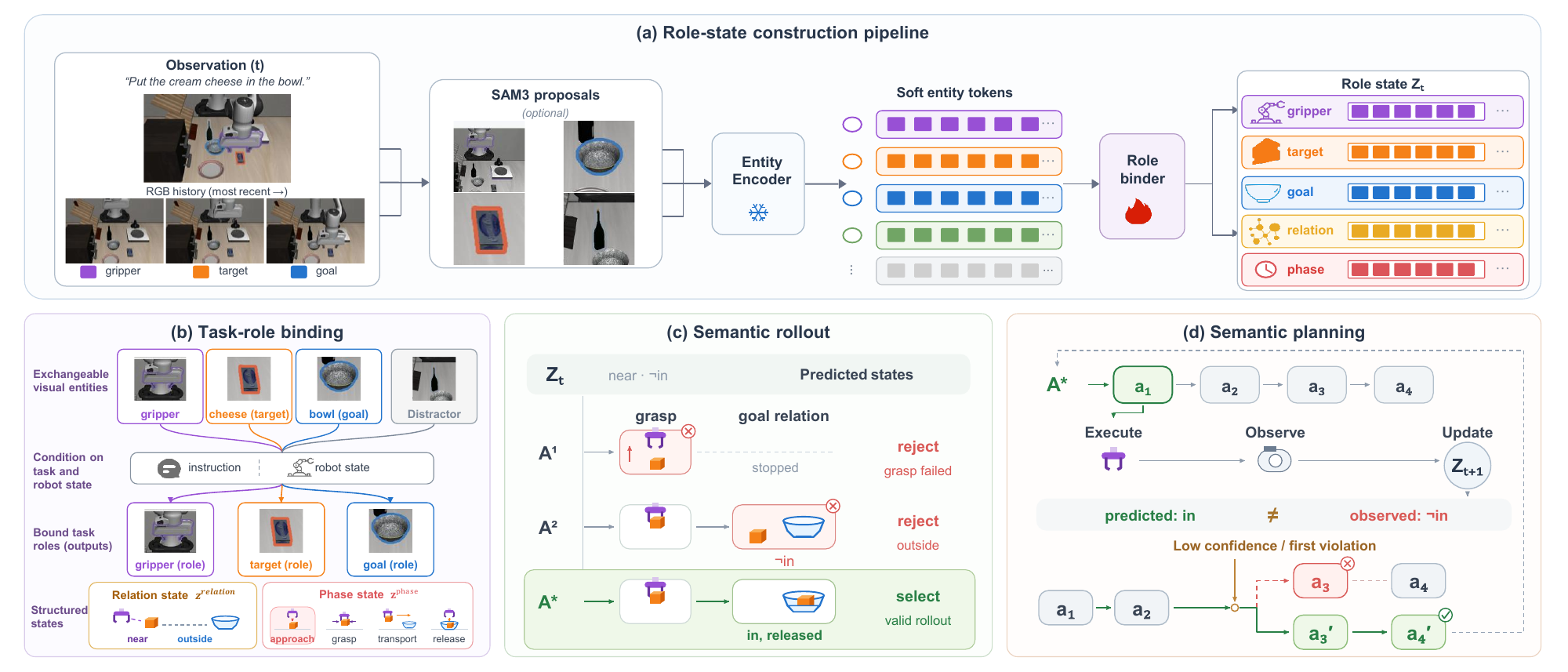}
\caption{\textbf{SemRole connects task grounding to action decisions.} \textbf{(a)} Visual features form soft entities, optionally aided by segmentation proposals. \textbf{(b)} Participant bindings, relations, and execution phase jointly form the semantic-role state. \textbf{(c)} Candidate rollouts predict grasp and goal-relation transitions for semantic selection. \textbf{(d)} The controller executes a prefix, updates the state from observations, and replaces a violating suffix when needed. Encoder status is schematic (Appendix~\ref{app:protocol-backbones}).}
\label{fig:overview}
\end{figure*}

\section{Semantic-Role-Structured World Models}
\label{sec:method}

Our objective is to make physical predictions useful for task-directed actions. \method{} constructs a joint role state that couples participant-specific information, relational conditions, and execution phase. We ground this state in observations, predict its changes under candidate actions, and use these changes to guide generation, selection, and local repair. Sections~\ref{sec:binding}--\ref{sec:repair} develop this shared interface.

\paragraph{Setting and notation.}
At control step $t\geq0$, the robot receives a finite history $o_t$ of RGB observations from $V$ cameras, a corresponding proprioceptive history $q_{\leq t}$, and an instruction $\ell$. A normalized action $a_t\in\mathbb{R}^{d_a}$ has dimension $d_a$. A candidate action chunk $A_t=(a_t,\ldots,a_{t+H-1})\in\mathbb{R}^{H\times d_a}$ spans $H$ control steps. The robot executes its first $h_{\mathrm{exec}}$ steps, with $1\leq h_{\mathrm{exec}}\leq H$, before observing again. We sample $K$ candidates per replan. For intermediate prediction, integer boundaries $0=h_0<h_1<\cdots<h_J=H$ divide each chunk into $J$ segments. Segment $j$ contains action indices $h_{j-1},\ldots,h_j-1$, relative to $t$. Parameters $\phi,\theta,\psi,\omega$ respectively denote the state-encoding, dynamics, action-generation, and selection networks.

\subsection{Grounding the Semantic-Role State in Observations}
\label{sec:binding}

\paragraph{Why object identity is insufficient.}
The encoder should reuse its description of a scene across instructions, while the controller must distinguish the object to manipulate from its destination. Exchangeable slot indices have no intrinsic connection to these functions. We therefore bind visual entities to task participants, then couple the resulting states to relation and phase information. This constructs the joint role state used by generation and prediction even when the relevant object identities change.

The visual backbone encodes the observation history into patch features. Learned queries aggregate these features into $N$ soft entity vectors $e_{t,n}$. Spatial coordinates, view identity, and temporal information support geometry and motion reasoning; soft attention permits an entity to collect evidence across views or partial occlusion. A segmentation proposal can bias this attention, but the full patch context remains available. Proposal dropout trains the same interface with and without that prior (Appendix~\ref{app:entities}).

Let $\mathcal{R}_{\mathrm{ent}}=\{\mathrm{grip},\mathrm{tgt},\mathrm{goal}\}$ index three participant-specific components: gripper, manipulated target, and goal. We use scaled dot-product attention \citep{vaswani2017attention}: for component $r$, a query $Q_r(\ell,q_t)\in\mathbb{R}^{d_k}$ attends to entity keys $K_e(e_{t,n})\in\mathbb{R}^{d_k}$, where $d_k$ is the attention-key dimension. A geometry descriptor $g_{t,n}\in\mathbb{R}^{d_g}$ and learned vector $w_r\in\mathbb{R}^{d_g}$ provide a component-specific spatial bias. The binding weights $b_{t,r,n}$ and the resulting $d$-dimensional component state are:
\begin{equation}
\begin{aligned}
b_{t,r,n}&=\operatorname{softmax}_{n}\!\Bigl(\rule{0pt}{1.4ex}\\
&\quad\frac{Q_r(\ell,q_t)^\top K_e(e_{t,n})}{\sqrt{d_k}}
+w_r^\top g_{t,n}\Bigr),\\
z_t^r&=U_r\!\left(\sum_{n=1}^{N}b_{t,r,n}V_e(e_{t,n}),\ell,q_t\right),\\
&\hspace{1em}r\in\mathcal{R}_{\mathrm{ent}}.
\end{aligned}
\label{eq:binding}
\end{equation}
Here $V_e$ projects values to $\mathbb{R}^d$, and $U_r$ fuses the attended evidence with the component's conditioning inputs. The gripper state can use proprioception directly, and the goal state can combine a fixture's appearance with its desired state, such as \emph{closed}. Binding is soft and need not be one-to-one: target and goal may refer to the same articulated fixture. The mechanism thus grounds task participation without requiring a persistent segmentation ID.

\begin{figure}[t]
\centering
\includegraphics[width=\linewidth]{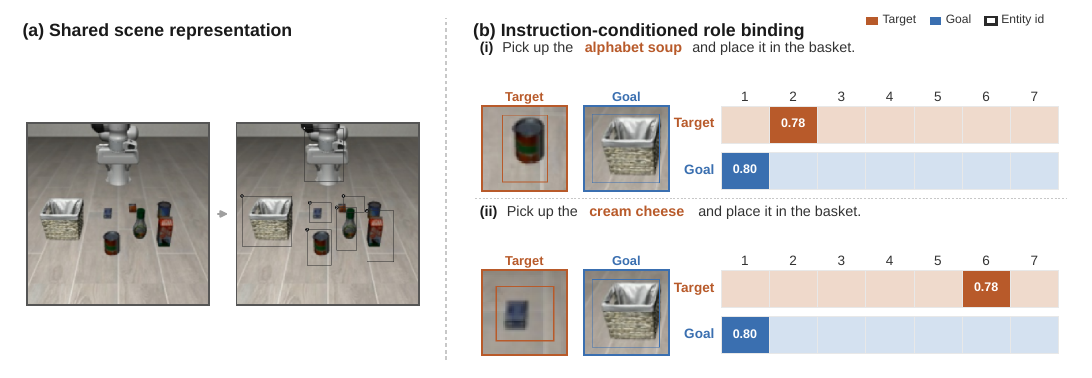}
\caption{\textbf{Task participation depends on the instruction.} The same scene supports different target assignments while the basket remains the goal. Target and goal bindings ground the participant components of the joint role state. The attention weights illustrate soft binding; quantitative grounding results appear in Table~\ref{tab:representation}.}
\label{fig:binding}
\end{figure}

\paragraph{Coupling participants to interaction context.}
Participant bindings supply only one aspect of semantic-role structure. Relations describe configurations and conditions involving those participants; execution phase records how the interaction has unfolded. A relation encoder $R_\phi$ combines the participant states with language, and a history encoder $P_\phi$ summarizes their evolution over $L$ observed steps, truncated at episode start. Projected to width $d$, the three participant components and two context components form one joint role state:
\begin{equation}
\begin{aligned}
z_t^{\mathrm{rel}}&=R_\phi(z_t^{\mathrm{grip}},z_t^{\mathrm{tgt}},z_t^{\mathrm{goal}},\ell),\\
z_t^{\mathrm{phase}}&=P_\phi\!\Bigl(\\
&\quad\{z_s^{\mathrm{grip}},z_s^{\mathrm{tgt}},z_s^{\mathrm{goal}},z_s^{\mathrm{rel}}\}
_{s=\max(0,t-L+1)}^{t},\ell\Bigr),\\
Z_t&=(z_t^{\mathrm{grip}},z_t^{\mathrm{tgt}},z_t^{\mathrm{goal}},
z_t^{\mathrm{rel}},z_t^{\mathrm{phase}})\\
&\quad\in\mathbb{R}^{5\times d}.
\end{aligned}
\label{eq:state}
\end{equation}
The five entries implement the joint representation; they are not five independent entity assignments. The relation component encodes geometry and predicates such as \emph{grasped}, \emph{inside}, or \emph{open}, while the phase component represents progress through the applicable task sequence. Entity bindings, predicate labels, and phase labels supervise this state during training without becoming privileged deployment inputs. Figure~\ref{fig:binding} illustrates how an instruction changes the target binding while preserving the scene representation.

\subsection{Predicting Transitions for Phase-Appropriate Progress}
\label{sec:dynamics}

\paragraph{Why an endpoint prediction is insufficient.}
A predicted final placement can conceal a lost grasp earlier in the rollout. Conversely, requiring every currently true predicate to remain true would incorrectly reject a necessary release. We therefore predict intermediate transitions and distinguish \emph{conditions to establish} from \emph{conditions still required at the current phase}.

For candidate $k$, let $A_{t,j}^k$ denote segment $j$ and initialize $\widehat Z_{t,0}^k=Z_t$. An action encoder $E_a$ preserves action order within each segment. A transformer $D_\theta$ exchanges information among role-state components and cross-attends to these action tokens. For every component $r\in\mathcal{R}=\mathcal{R}_{\mathrm{ent}}\cup\{\mathrm{rel},\mathrm{phase}\}$, it predicts a residual $\Delta z_{t,j}^{r,k}\in\mathbb{R}^d$ and a gate logit $\gamma_{t,j}^{r,k}\in\mathbb{R}^d$. The recurrent update is:
\begin{equation}
\begin{aligned}
&(\Delta z_{t,j}^{r,k},\gamma_{t,j}^{r,k})\\
&\quad=D_\theta^r(\widehat Z_{t,j-1}^k,E_a(A_{t,j}^k),\ell),\\
&\widehat z_{t,j}^{r,k}=\widehat z_{t,j-1}^{r,k}
+\sigma(\gamma_{t,j}^{r,k})\odot\Delta z_{t,j}^{r,k}.
\end{aligned}
\label{eq:dynamics}
\end{equation}
Here $\sigma$ is the elementwise sigmoid and $\odot$ denotes elementwise multiplication. State $\widehat Z_{t,j}^k$ predicts the scene at time $t+h_j$. The gate allows unaffected components to change little while affected components change substantially; it does not itself impose a physical conservation law.

\paragraph{Separating establishment from preservation.}
Semantic heads read consecutive predicted states and the intervening action segment. They predict contact and grasp status, active predicate establishment, preservation of required predicates, conjunction satisfaction, fixture state, and the next phase and transition validity. For example, transport requires a stable grasp; release intentionally ends that requirement and should instead establish supported placement. Preservation labels therefore apply only while a condition remains required by the task phase. This prevents the model from treating deliberate release as grasp failure. Appendix~\ref{app:supervision} defines the corresponding label masks and losses.

In simulation, predicates and fixture states are derived from task specifications and simulator state. Contact and grasp supervision additionally use contact, gripper, and relative-motion information. These signals supervise predictions from observed role states, rather than supplying future state to the deployed planner. Unavailable labels are masked out. Predicting at $J$ boundaries localizes a suspected violation to an action segment; it does not claim to identify an exact continuous-time contact event.

\paragraph{Making action order matter.}
Recognizing the task from the current image is not enough to compare two actions from that image. We train with temporal shuffles, reversed segments, and task-matched substitutions to construct hard negatives. Such a transformation is a negative only when its label indicates reduced task validity; shuffling is not intrinsically a failure. Let $A^+$ and $A^-$ be a valid chunk and a labeled negative from the same context, with semantic scores $S^+$ and $S^-$. For margin $m>0$, the ranking objective is:
\begin{equation}
\mathcal L_{\mathrm{rank}}=\mathbb E_{(A^+,A^-)}
\left[\max(0,m-S^++S^-)\right].
\label{eq:rank}
\end{equation}
This loss encourages action-sensitive scores. Its evaluation tests discrimination of the constructed pairs, rather than proving causal identification of the environment dynamics.

\subsection{Generating and Selecting Task-Consistent Actions}
\label{sec:selection}

\paragraph{Why semantics must enter before selection.}
A selector cannot recover an action absent from its candidate set. We therefore condition action generation on the same $Z_t$ used for prediction, so that target binding, geometry, and phase influence which candidates are proposed. A conditional flow-matching head \citep{lipman2023flow} learns a vector field $v_\psi$ over normalized action chunks. Given a demonstrated chunk $A^+$, standard Gaussian noise $\epsilon$ of shape $H\times d_a$, and flow time $\tau\sim\mathcal U[0,1]$, the training path and objective are:
\begin{equation}
\begin{aligned}
&A^\tau=(1-\tau)\epsilon+\tau A^+,\\
&\mathcal L_{\mathrm{FM}}=\mathbb E_{A^+,Z_t,\ell,\epsilon,\tau}\!\Biggl[\\
&\qquad\frac{\|v_\psi(A^\tau,\tau\mid Z_t,\ell)-(A^+-\epsilon)\|_F^2}{H d_a}
\Biggr].
\end{aligned}
\label{eq:fm}
\end{equation}
The squared Frobenius norm $\|\cdot\|_F^2$ sums squared errors over time and action coordinates. Flow time $\tau$ is distinct from robot time $t$. At inference, integrating $dA^\tau/d\tau=v_\psi(A^\tau,\tau\mid Z_t,\ell)$ from $A^0=\epsilon$ to $\tau=1$ gives a candidate $A_t^k$. All $K$ candidates share weights and differ only in independently sampled noise.

\paragraph{Scoring predicted effects in context.}
Candidates with similar endpoints can differ in whether they violate a grasp or fixture requirement along the way. We score the predicted transitions in their phase context. Let $\widehat Y_{t,j}^k$ collect the semantic predictions at segment $j$, including probabilities for binary conditions and the categorical phase distribution. Let $u_t^k$ summarize their predictive uncertainty. The selector $s_\omega$ receives the current phase and the ordered transition predictions, producing one score per candidate:
\begin{equation}
\begin{aligned}
S_t^k&=s_\omega\!\left(z_t^{\mathrm{phase}},
(\widehat Y_{t,1}^k,\ldots,\widehat Y_{t,J}^k),u_t^k\right),\\
k^\star&=\arg\max_{1\leq k\leq K}S_t^k.
\end{aligned}
\label{eq:score}
\end{equation}
The phase context lets the same event receive different significance during approach and release. Intermediate establishment, preservation, and transition-validity predictions expose failures hidden by endpoint agreement. The score is a learned ranking quantity, not a calibrated probability of success. Predictive uncertainty is a diagnostic input and cannot certify that an action is safe.

\subsection{Repairing the First Invalid Suffix}
\label{sec:repair}

\paragraph{Why restarting a good prefix is wasteful.}
A candidate may correctly approach and grasp a target, then release before reaching the goal. Full resampling replaces all three decisions. \method{} instead retains the prefix preceding the earliest predicted violation and generates a new suffix conditioned on it. Let $j^\star$ be the first segment whose semantic predictions violate an applicable requirement, and set the retained prefix length to $b=h_{j^\star-1}$. If the best score is below a validation-set threshold $\eta$ without a localized violation, use $b=0$: low confidence alone provides no evidence that any prefix is valid.

Let $M_b\in\{0,1\}^{H\times d_a}$ equal one in the first $b$ action rows and zero elsewhere, and let $\bar M_b=1-M_b$. With the selected chunk $A_t^{k^\star}$ as the anchor, prefix-conditioned flow generation is:
\begin{equation}
\begin{aligned}
\widetilde A^0&=M_b\odot A_t^{k^\star}+\bar M_b\odot\epsilon,\\
\frac{d\widetilde A^\tau}{d\tau}
&=\bar M_b\odot v_\psi\!\Bigl(\\
&\quad\widetilde A^\tau,\tau\mid Z_t,\ell,M_b,M_b\odot A_t^{k^\star}\Bigr).
\end{aligned}
\label{eq:repair}
\end{equation}
The masked derivative keeps every prefix coordinate constant. This follows the trajectory-inpainting view of conditional planning \citep{janner2022planning}, with the repair boundary chosen from predicted semantic violations. Training freezes random prefixes and supervises suffix velocity (Appendix~\ref{app:repair}). Repaired candidates are rolled out and scored again; retaining a prefix does not exempt it from re-evaluation. An empty prefix reduces to full resampling. Repairs use a bounded proposal budget, after which the controller executes a short accepted prefix or invokes its rejection fallback. New observations reconstruct $Z_t$ before the next decision, limiting dependence on an imagined trajectory.

\paragraph{Learning and backbone interface.}
The same typed state connects the visual encoder, action generator, dynamics, and selector. We retain backbone representation learning and combine it with semantic supervision, action generation, action ranking, and masked repair. For nonnegative loss weights $\lambda_{\mathrm{rep}},\lambda_{\mathrm{sem}},\lambda_{\mathrm{FM}},\lambda_{\mathrm{rank}},\lambda_{\mathrm{repair}}$, the objective is:
\begin{equation}
\begin{aligned}
\mathcal L&=\lambda_{\mathrm{rep}}\mathcal L_{\mathrm{rep}}
+\lambda_{\mathrm{sem}}\mathcal L_{\mathrm{sem}}\\
&\quad+\lambda_{\mathrm{FM}}\mathcal L_{\mathrm{FM}}
+\lambda_{\mathrm{rank}}\mathcal L_{\mathrm{rank}}\\
&\quad+\lambda_{\mathrm{repair}}\mathcal L_{\mathrm{repair}}.
\end{aligned}
\label{eq:total}
\end{equation}
Appendix~\ref{app:training} specifies the loss interfaces and distinguishes them from implementation settings still requiring documentation. We instantiate the visual interface with \backbone{} on LIBERO and pretrained \cosmos{} features on RoboCasa and Galbot. The two lines share the role-state structure and decision procedure; comparisons are made within each backbone. A compact backbone does not imply that the complete system, including its action head, has the same parameter count.

%% file: sections/04_experiments.tex
\section{Experiments}
\label{sec:experiments}

We ask: \textbf{(Q1)} does role structure improve candidates before selection; \textbf{(Q2)} do semantic selection and local repair convert coverage into decisions; and \textbf{(Q3)} does the design transfer across backbones, tasks, and physical execution?

\paragraph{Settings and comparisons.}
\backbone{}: LIBERO-Object/Spatial/Goal/Long \citep{liu2023libero}. \cosmos{}: 17 RoboCasa atomic tasks \citep{nasiriany2024robocasa} and Galbot. Matched-backbone baselines use a flow-matching prior on backbone features ($K{=}4$), not the unmodified published backbone scores.

The primary metric is closed-loop task success. LIBERO and RoboCasa use 50 trials per task for each of three training seeds. We weight the four LIBERO suites equally and macro-average the 17 RoboCasa tasks. Full resampling and suffix repair share a budget of at most three flow integrations per decision. Appendix~\ref{app:protocol} gives the full evaluation protocol.

\subsection{Role Structure Improves Both Proposals and Decisions}
\label{sec:libero}

\paragraph{Q1--Q2: Proposals, selection, and repair.}
Role conditioning lifts Candidate~0 from 45.5\% to 62.8\% before any learned selector (Table~\ref{tab:libero}). Semantic selection reaches 75.3\%. With the same repair budget, suffix repair reaches 83.0\% compared with 78.5\% for full resampling. Long remains the hardest suite, at 72\% with suffix repair.

\input{tables/libero}

\paragraph{Candidate controls.}
Table~\ref{tab:libero-diagnostics} separates proposal coverage from deployable selection. With the role-conditioned prior, random selection reaches 61.8\% and the fixed candidate index selected on validation reaches 64.8\%, compared with 75.3\% for semantic selection (Table~\ref{tab:libero}). The monolithic and instance-slot selectors reach 50.5\% and 55.8\%. Oracle@4 selects using simulator chunk outcomes at each replan. It reaches 68.0\% with the latent prior and 89.5\% with the role-conditioned prior. These oracle scores apply to the original candidate pool, which suffix repair expands.

\input{tables/libero_diagnostics}

\subsection{The Same Task Structure Helps a Foundation Backbone}
\label{sec:robocasa}

\paragraph{Q3: Kitchen tasks with Cosmos.}
Table~\ref{tab:robocasa} evaluates cumulative additions to the \cosmos{} action prior. Role conditioning improves the mean from 42.9\% to 45.2\%; semantic scoring raises it to 51.8\%; suffix repair reaches 55.2\%. The 12.3-point overall gain is smaller than on LIBERO, and most of it appears after semantic scoring is added. This difference cautions against attributing one uniform mechanism of improvement to every backbone and task distribution.

Hard-subset mean rises 29.3\%$\to$46.1\% (largest gains: stove off / cabinet / microwave on); two tasks regress slightly (Appendix~\ref{app:robocasa}).

\input{tables/robocasa_summary}

\paragraph{Physical execution on Galbot.}
We evaluate tabletop placement and microwave retrieval using the \cosmos{} backbone. Starting from simulation pretraining, the robot adaptation uses 20 teleoperated demonstrations per task and 2k updates. We evaluate one adapted checkpoint, with $K{=}4$ and 50 independent, reset-matched trials per task and method. The 60\,s timeout, operator intervention, and e-stop all count toward failure (Appendix~\ref{app:galbot}).

\begin{figure*}[t]
\centering
\includegraphics[width=\linewidth,keepaspectratio]{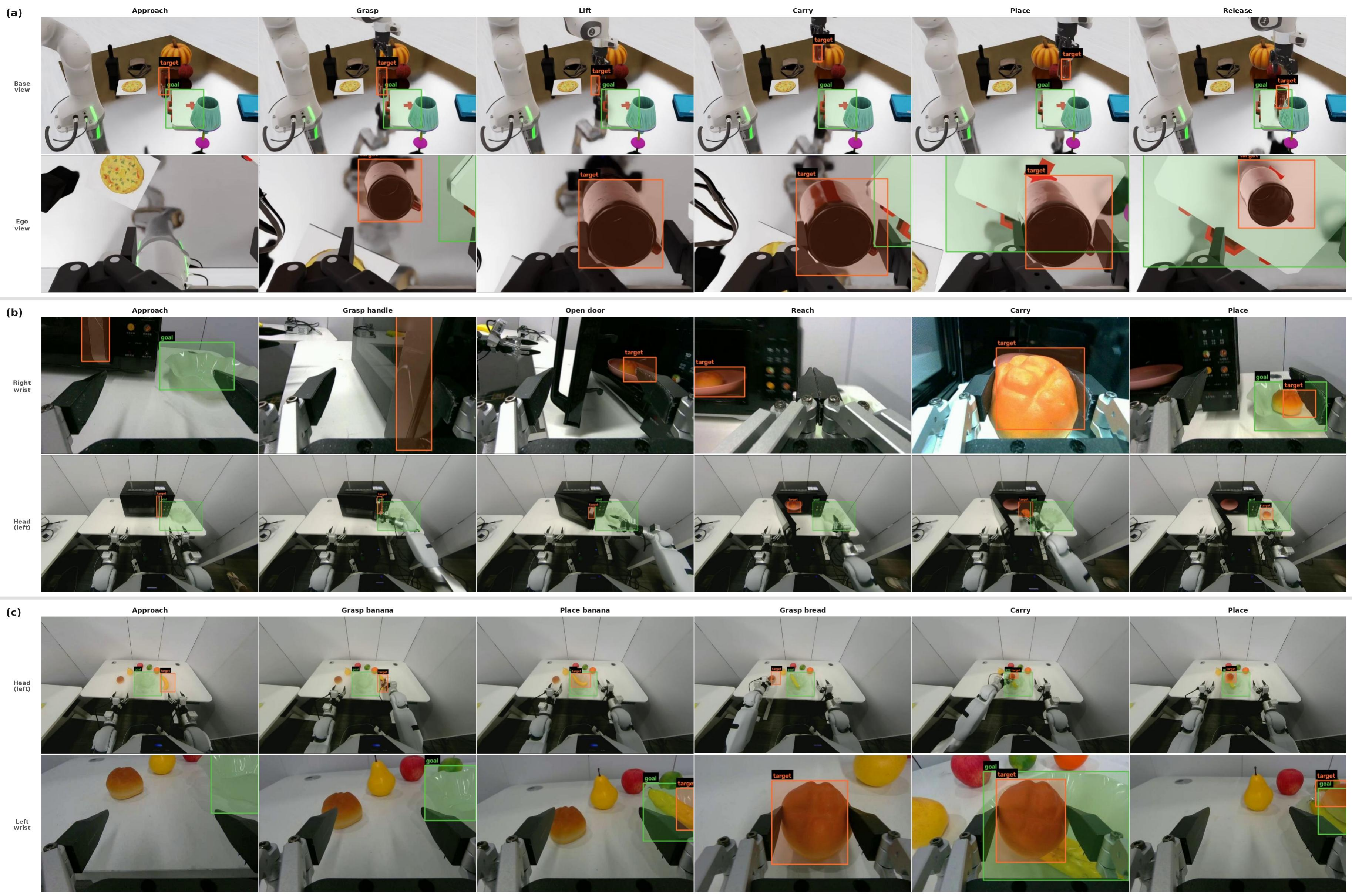}
\caption{\textbf{Qualitative rollouts in simulation and on the robot.} (a) RoboCasa mug placement. (b,c) Galbot retrieval and plate placement. Orange and green boxes respectively mark Target and Goal. Table~\ref{tab:galbot} reports the physical success rates.}
\label{fig:galbot}
\end{figure*}

Figure~\ref{fig:galbot} shows example retrieval and placement sequences.

Success improves on both tasks, with a 26.0-point gain in the unweighted mean (Table~\ref{tab:galbot}).

\input{tables/galbot}

\subsection{Which Structure and Predictions Matter?}
\label{sec:analysis}

\paragraph{Role structure.}
Table~\ref{tab:roles-main} traces the stages of a separate representation ablation. Adding instruction to object slots lifts success from 55.0\% to 62.0\%. Explicit target and goal roles reach 68.0\%, and adding relations reaches 73.0\%. The full state reaches 83.0\% after adding gripper and phase jointly. The 55.0\% starting point belongs to this ablation. The instance-slot selector control in Table~\ref{tab:libero-diagnostics} is a separate comparison.

\input{tables/roles_main}

\paragraph{Prediction vs.\ decisions.}
The full model ranks the valid action above the labeled negative in 78.2\% of pairs. Removing shuffled negatives lowers ranking accuracy to 53.7\%. Removing phase loss improves role-prediction error while ranking falls to 66.8\%. The two metrics favor different variants (Table~\ref{tab:action-sensitivity}).

\input{tables/action_sensitivity}

\paragraph{Transfer and perceptual dependence.}
With target data and updates matched, Object pretraining improves success over scratch by 8.3 points on Goal and 8.1 on Long. On \mbox{Spatial}, adaptation reaches 72.1\% versus 72.8\% from scratch (Table~\ref{tab:cross-suite}).

In the perception diagnostic, success is 72.6\% without masks, 75.1\% with a soft segmentation prior, and 70.8\% with a corrupted prior (Table~\ref{tab:representation}).

\input{tables/efficiency}

\paragraph{Component costs.}
The full \backbone{}-based system has 45.4M parameters. At $K{=}4$, generation and scoring take 30.2 and 7.3\,ms per replan, respectively (Table~\ref{tab:efficiency}). Under the same budget, the latent baseline takes 28.7 and 4.8\,ms. These timings isolate the components and exclude repeated repair.

%% file: tables/libero.tex
\begin{table}[htbp]
\centering
\caption{\textbf{Generation, selection, and repair on LIBERO.} Success (\%) with the \backbone{} backbone. Avg.\ equally weights the four suites. Candidate~0 uses the first sample without learned selection. Table~\ref{tab:libero-diagnostics} reports the representation and candidate controls.}
\label{tab:libero}
\tabstyle
\setlength{\tabcolsep}{2pt}
\begin{tabularx}{\linewidth}{@{}>{\raggedright\arraybackslash}Xrrrrr@{}}
\toprule
State / decision rule & Object & Spatial & Goal & Long & Avg.\\
\midrule
LeWM latent: Candidate~0 & 58 & 50 & 46 & 28 & 45.5\\
SemRole: Candidate~0 & 74 & 67 & 63 & 47 & 62.8\\
\quad + semantic selector & 85 & 79 & 75 & 62 & 75.3\\
\quad + full resampling & 88 & 82 & 78 & 66 & 78.5\\
\quad + suffix repair & \textbf{91} & \textbf{86} & \textbf{83} & \textbf{72} & \textbf{83.0}\\
\bottomrule
\end{tabularx}
\end{table}

%% file: tables/libero_diagnostics.tex
\begin{table}[htbp]
\centering
\caption{\textbf{LIBERO representation and candidate controls.} Success (\%) under the protocol in Table~\ref{tab:libero}. Random@4 samples uniformly from four proposals. The best fixed ID is selected on validation and held fixed at test time. Oracle@4 uses simulator outcomes for the original candidates at each replan.}
\label{tab:libero-diagnostics}
\tabstyle
\setlength{\tabcolsep}{2pt}
\begin{tabularx}{\linewidth}{@{}>{\raggedright\arraybackslash}Xrrrrr@{}}
\toprule
State / decision rule & Object & Spatial & Goal & Long & Avg.\\
\midrule
LeWM latent: monolithic selector & 63 & 55 & 51 & 33 & 50.5\\
Instance slots: slot selector & 68 & 60 & 56 & 39 & 55.8\\
SemRole: Random@4 & 73 & 66 & 62 & 46 & 61.8\\
SemRole: best fixed ID (val) & 76 & 69 & 65 & 49 & 64.8\\
LeWM latent: Oracle@4 & 80 & 73 & 68 & 51 & 68.0\\
SemRole: Oracle@4 & 96 & 92 & 89 & 81 & 89.5\\
\bottomrule
\end{tabularx}
\end{table}

%% file: tables/robocasa_summary.tex
\begin{table}[htbp]
\centering
\scriptsize
\caption{\textbf{Cumulative components on RoboCasa with Cosmos.} Macro-average success (\%) over 17 tasks. +Role changes generation; +Score adds semantic selection; +Repair adds suffix repair. The hard subset is a descriptive grouping of 11 tasks with Base $\leq45\%$, not an independent benchmark split. $\Delta$ is +Repair minus Base in percentage points.}
\label{tab:robocasa}
\tabstyle
\setlength{\tabcolsep}{2pt}
\begin{tabularx}{\linewidth}{@{}>{\raggedright\arraybackslash}Xrrrrrr@{}}
\toprule
Tasks & Base & +Role & +Score & +Repair & Oracle@4 & $\Delta$\\
\midrule
All 17 & 42.9 & 45.2 & 51.8 & \textbf{55.2} & 61.4 & +12.3\\
Hard 11 & 29.3 & 32.4 & 41.4 & \textbf{46.1} & 54.8 & +16.8\\
\bottomrule
\end{tabularx}
\end{table}

%% file: tables/galbot.tex
\begin{table}[htbp]
\centering
\caption{\textbf{Galbot physical evaluation with Cosmos.} Closed-loop success (\%) over 50 independent, reset-matched trials per task and method. Avg.\ weights both tasks equally.}
\label{tab:galbot}
\tabstyle
\begin{tabular*}{\linewidth}{@{\extracolsep{\fill}}lrrr@{}}
\toprule
Method & Tabletop & Microwave & Avg. \\
\midrule
Cosmos baseline & 68.0 & 42.0 & 55.0 \\
Cosmos + \method{} & \textbf{88.0} & \textbf{74.0} & \textbf{81.0} \\
\bottomrule
\end{tabular*}
\end{table}

%% file: tables/roles_main.tex
\begin{table}[htbp]
\centering
\caption{\textbf{Task structure in the state representation.} LIBERO success (\%) for progressively structured states. Reported $\pm$ is mean $\pm$ std over training seeds $\{0,1,2\}$, each with 50 trials/task (Appendix~\ref{app:protocol}). The last row jointly introduces gripper and phase, so it does not isolate their individual effects.}
\label{tab:roles-main}
\tabstyle
\begin{tabular*}{\linewidth}{@{\extracolsep{\fill}}lr@{}}
\toprule
Representation & Success (\%)\\
\midrule
Five instance slots & $55.0\pm3.6$\\
Object slots + instruction & $62.0\pm3.3$\\
Target + goal roles & $68.0\pm3.1$\\
Target + goal + relation & $73.0\pm2.9$\\
SemRole (+ gripper, phase) & $\mathbf{83.0\pm2.4}$\\
\bottomrule
\end{tabular*}
\end{table}

%% file: tables/action_sensitivity.tex
\begin{table}[t]
\centering
\scriptsize
\caption{\textbf{Action-sensitivity diagnostics.}
Role error is future role-state MSE (mean $\pm$ SD across three training seeds).
Gap is the mean positive-minus-negative score. Rank accuracy is in percent.}
\label{tab:action-sensitivity}
\tabstyle
\setlength{\tabcolsep}{2pt}
\begin{tabularx}{\linewidth}{@{}>{\raggedright\arraybackslash}Xrrr@{}}
\toprule
\textbf{Variant} & Role error $\downarrow$ & Gap $\uparrow$ & Rank acc. $\uparrow$ \\
\midrule
No semantic heads & $0.176{\pm}0.011$ & $0.061{\pm}0.019$ & $60.8{\pm}2.9$ \\
No shuffled negatives & $0.149{\pm}0.009$ & $0.018{\pm}0.011$ & $53.7{\pm}3.3$ \\
No preservation loss & $0.158{\pm}0.010$ & $0.108{\pm}0.023$ & $69.4{\pm}2.6$ \\
No phase loss & $0.141{\pm}0.008$ & $0.087{\pm}0.020$ & $66.8{\pm}2.8$ \\
\method{} & $0.144{\pm}0.007$ & $0.157{\pm}0.025$ & $78.2{\pm}2.1$ \\
\bottomrule
\end{tabularx}
\end{table}

%% file: tables/efficiency.tex
\begin{table}[t]
\centering
\caption{\textbf{Complete-system size and component costs.}
(a) Total and trainable parameters, and peak training memory in GB.
(b) Generation and scoring in ms per replan at $K{=}4$.
$^\dagger$Published backbone size. Timing protocol and exclusions are in Appendix~\ref{app:efficiency}.}
\label{tab:efficiency}
\tabstyle
\setlength{\tabcolsep}{2pt}
\begin{tabularx}{\linewidth}{@{}>{\raggedright\arraybackslash}Xrrr@{}}
\toprule
\multicolumn{4}{@{}l}{\textbf{(a) Parameters and peak memory}}\\
\midrule
\textbf{Model} & \textbf{Total} & \textbf{Trainable} & \textbf{Memory}\\
\midrule
\backbone{} & 15.0M$^\dagger$ & 15.0M & $4.3{\pm}0.2$\\
Latent + FM & 35.8M & 35.8M & $8.1{\pm}0.3$\\
Instance slots + FM & 39.1M & 39.1M & $8.9{\pm}0.4$\\
\method{} (world only) & 24.6M & 24.6M & $6.6{\pm}0.3$\\
\textbf{\method{} + FM} & 45.4M & 45.4M & $9.2{\pm}0.4$\\
\bottomrule
\end{tabularx}
\par\smallskip
\begin{tabularx}{\linewidth}{@{}>{\raggedright\arraybackslash}Xrr@{}}
\toprule
\multicolumn{3}{@{}l}{\textbf{(b) Generation and scoring latency}}\\
\midrule
\textbf{Model} & \textbf{Gen.} & \textbf{Score}\\
\midrule
\backbone{} & n/a & $4.4{\pm}0.3$\\
Latent + FM & $28.7{\pm}1.1$ & $4.8{\pm}0.3$\\
Instance slots + FM & $31.0{\pm}1.4$ & $7.6{\pm}0.6$\\
\method{} (world only) & n/a & $7.1{\pm}0.5$\\
\textbf{\method{} + FM} & $30.2{\pm}1.2$ & $7.3{\pm}0.5$\\
\bottomrule
\end{tabularx}
\end{table}

%% file: sections/05_conclusion.tex
\section{Limitations and Conclusion}
\label{sec:conclusion}

The five-component role state assumes a dominant target and goal. Multiple targets, bimanual coordination, deformable objects, and tool use may require richer role structures. Semantic labels add supervision costs, especially outside simulation. Predictions can miss brief contacts and accept physically invalid prefixes. Suffix repair adds computation beyond the reported component timings.

\method{} uses participant roles, relations, and execution phase to guide action generation, selection, and repair. Sharing this state improves closed-loop success on LIBERO, RoboCasa, and Galbot. The results show the value of a predictive state that distinguishes when a condition must be preserved from when it should change.

%% file: sections/06_statements.tex
\section*{LLM Usage Statement}
\label{llmstatementstart}
OpenAI's ChatGPT (GPT models) and Codex were used to assist with translation and language editing of author-written text, provide reviewer-style feedback, and refine author-prepared descriptions of the experiments and data. Codex also assisted with checking bibliographic information, adding a citation to the related-work section, and editing and debugging the LaTeX source and layout-checking scripts. ChatGPT's image-generation tools were used to create selected icons for the figures. The reported experimental measurements were supplied by the authors. The authors take full responsibility for the submitted content, including material prepared with AI assistance.

%% file: supplementary.tex
\clearpage
\raggedbottom
\setcounter{page}{1}
\renewcommand{\thepage}{S\arabic{page}}
\maketitlesupplementary
\phantomsection
\label{supplementarystart}
\appendix
\input{sections/A_specification}
\input{sections/B_results}
\input{sections/C_protocol}
\input{sections/D_reproducibility}

%% file: sections/A_specification.tex
\section{Mathematical Specification and Training Interfaces}
\label{app:specification}

This appendix makes the proposed method explicit where the source description left interfaces implicit. In particular, it specifies phase-dependent label masks and prefix-conditioned generation. The corresponding implementation choices and hyperparameters must be checked against the experimental system; fields not recoverable from the supplied experiment summaries are marked in Appendix~\ref{app:protocol}.

\subsection{Notation and Typed State}
\begin{table}[htb]
\centering
\caption{Symbols used in the method. Robot time, flow time, and rollout-segment indices are distinct.}
\label{tab:notation}
\tabstyle
\begin{tabularx}{\linewidth}{@{}lX@{}}
\toprule
Symbol & Meaning\\
\midrule
$t$, $\tau$, $j$ & Robot control step, continuous flow time, and rollout-segment index.\\
$H$, $d_a$, $K$ & Action horizon, action dimension, and initial candidate count.\\
$J$, $h_j$ & Number of segments and end boundary of segment $j$.\\
$L$, $V$, $N$ & Role-history length, camera count, and number of soft entities.\\
$d$, $d_k$, $d_g$ & State-component width, attention-key width, and geometry-feature width.\\
$e_{t,n}$, $b_{t,r,n}$ & Entity $n$ and its binding weight for participant component $r$.\\
$Z_t$, $\widehat Z_{t,j}^k$ & Observed role state and predicted state at $t+h_j$ for candidate $k$.\\
$A_t^k$, $A_{t,j}^k$ & Candidate action chunk and its $j$th action segment.\\
$\widehat Y_{t,j}^k$, $S_t^k$ & Predicted semantic transitions and candidate score.\\
$j^\star$, $b$, $M_b$ & First violating segment, retained action count, and prefix mask.\\
$h_{\mathrm{exec}}$, $R$ & Executed prefix length and maximum repair-proposal count.\\
\bottomrule
\end{tabularx}
\end{table}

The joint role state $Z_t$ is implemented by five components with fixed meanings. The participant components ground the end effector and grasp-relevant robot state (gripper), the object or articulated part to manipulate (target), and the destination or desired fixture state (goal). Their entity bindings can change with the instruction. The relation component combines relative geometry with task predicates; phase represents progress through applicable stages. Together they specify task participation in its interaction context. Tasks may have different stages, and a goal need not be a separate physical object.

\subsection{Soft Entities and Optional Segmentation Priors}
\label{app:entities}

Flattening the patch tokens across views and history gives tokens $x_i$, with $i=1,\ldots,P$ and $P$ the total patch count. For entity query $q_n^{\mathrm{ent}}$ and patch key $k_i^{\mathrm{patch}}$ in a shared $d_k$-dimensional space, let $M_{n,i}$ be a proposal affinity aligned to query $n$. With nonnegative bias strength $\lambda_{\mathrm{seg}}$, the entity attention is:
\begin{equation}
\alpha_{n,i}=\operatorname{softmax}_{i}\!\left(
\frac{(q_n^{\mathrm{ent}})^\top k_i^{\mathrm{patch}}}{\sqrt{d_k}}
+\lambda_{\mathrm{seg}}M_{n,i}\right).
\label{eq:entities}
\end{equation}
The entity vector aggregates projected patch values with these weights. Proposal masks influence attention, not the availability of image patches. Setting $\lambda_{\mathrm{seg}}=0$ recovers patch-only inference. Proposal-set dropout, individual-mask dropout, and bias-strength variation expose training to missing or unreliable priors. The proposal-to-query association and corruption distributions are implementation details that must be supplied; a raw segmentation ID cannot be assumed to equal a learned entity index. For the same reason, mask identity is not the definition of target or goal.

\paragraph{Entity permutation preserves participant binding.}
Consider a permutation $\pi$ of the entities and apply it consistently to $e_{t,n}$ and $g_{t,n}$. Because the key, value, and geometry functions are shared across entity indices, the attention logits in Eq.~\ref{eq:binding} are permuted by $\pi$. Softmax permutes in the same way. The weighted sum is unchanged after reindexing, so $z_t^r$ is invariant to this permutation. The argument requires that no additional entity-index embedding enters the binder. It is a property of the stated binding equation, not a claim that segmentation or temporal tracking is error-free.

\subsection{Semantic Labels, Applicability, and Missing Supervision}
\label{app:supervision}

Let $s_{t+h_j}$ denote the simulator or annotated physical state at boundary $j$. For predicate $p$ in a task's predicate set $\mathcal P$, let $y_{j,p}=f_p(s_{t+h_j},\ell)\in\{0,1\}$ be its ground-truth truth value. Examples include object containment, support, and fixture opening; $f_p$ is the corresponding task-specific labeling rule. Let $\rho_{j,p}\in\{0,1\}$ indicate whether predicate $p$ is required at boundary $j$ under the labeled phase. Establishment events and the mask for preservation supervision are:
\begin{equation}
\begin{aligned}
y_{j,p}^{\mathrm{est}}&=(1-y_{j-1,p})y_{j,p},\\
\mu_{j,p}&=\rho_{j-1,p}\rho_{j,p}y_{j-1,p}.
\end{aligned}
\label{eq:predicate-labels}
\end{equation}
The establishment target is $y_{j,p}^{\mathrm{est}}$ whenever this transition is relevant and observed. Preservation predicts $y_{j,p}$ only where $\mu_{j,p}=1$: the condition held before the segment and is required at both boundaries. Missing labels add a further availability mask. During an intentional release, the task no longer requires a grasp, so its preservation mask is zero. An accidentally lost grasp during transport still receives a negative preservation target. Requirements spanning an entire segment must be labeled consistently with intermediate events; endpoint labels alone cannot detect a brief violation followed by recovery.

For a nonempty set $\mathcal P_{\mathrm{goal}}$ of required terminal predicates, the conjunction target is defined from ground-truth labels rather than assuming independence between predicted probabilities:
\begin{equation}
y_j^{\mathrm{conj}}=\prod_{p\in\mathcal P_{\mathrm{goal}}}y_{j,p}.
\label{eq:conjunction}
\end{equation}
Contact and stable grasp are separate targets: proximity or momentary collision alone does not establish a stable grasp. Their labeling needs contact, aperture, relative-motion criteria, and temporal windows. A phase label $c_j\in\{1,\ldots,C\}$ identifies one of $C$ applicable stages. Transition validity is task-dependent and must allow all legal progress within a segment, including more than one adjacent phase change when the temporal resolution permits it. Continuous fixture targets, such as joint displacement or angle, require declared units and normalization. No numerical labeling thresholds are inferred from task names.

\paragraph{Normalized supervision.}
Let $\mathcal I_{\mathrm{bin}}$ be the set of valid binary targets across examples, segments, and binary heads. For target $y_i$ and predicted probability $\widehat p_i$, binary cross-entropy (BCE) is averaged over valid labels:
\begin{equation}
\begin{aligned}
\mathcal L_{\mathrm{bin}}&=
\frac{1}{\max(1,|\mathcal I_{\mathrm{bin}}|)}\\
&\quad\times\sum_{i\in\mathcal I_{\mathrm{bin}}}
\bigl[-y_i\log\widehat p_i\\
&\qquad\qquad-(1-y_i)\log(1-\widehat p_i)\bigr].
\end{aligned}
\label{eq:binary-loss}
\end{equation}
An empty sum contributes zero, and the denominator prevents division by zero when all labels are missing. Probabilities are numerically clipped away from zero and one when evaluating the logarithms. Binding uses cross-entropy against a soft assignment over entities derived from the annotated regions; phase uses categorical cross-entropy; continuous normalized fixture states use regression loss. With nonnegative weights $\beta_{\mathrm{bind}},\beta_{\mathrm{bin}},\beta_{\mathrm{phase}},\beta_{\mathrm{fixture}}$, these form:
\begin{equation}
\begin{aligned}
\mathcal L_{\mathrm{sem}}&=
\beta_{\mathrm{bind}}\mathcal L_{\mathrm{bind}}
+\beta_{\mathrm{bin}}\mathcal L_{\mathrm{bin}}\\
&\quad+\beta_{\mathrm{phase}}\mathcal L_{\mathrm{phase}}
+\beta_{\mathrm{fixture}}\mathcal L_{\mathrm{fixture}}.
\end{aligned}
\label{eq:sem-loss}
\end{equation}
Losses average only over applicable, observed targets. Contact, grasp, establishment, preservation, conjunction, and transition-validity heads are included in $\mathcal L_{\mathrm{bin}}$; they need not share a label mask. Binding targets for broad destinations may be distributions rather than one-hot object identities. The annotation-to-entity matching rule must respect the encoder's current entity ordering; a raw object ID is not an entity-slot index.

\subsection{Prefix-Conditioned Flow Matching}
\label{app:repair}

For repair training, choose a demonstrated action chunk $A^+$ and a prefix length $b\in\{0,\ldots,H-1\}$. The mask $M_b$ fixes all action coordinates before $b$. Its complement $\bar M_b$ contains exactly $(H-b)d_a$ coordinates. The conditional path $A_b^\tau$ and repair objective are:
\begin{equation}
\begin{aligned}
&A_b^\tau=M_b\odot A^+\\
&\qquad+\bar M_b\odot[(1-\tau)\epsilon+\tau A^+],\\
&\widehat v_b^\tau=v_\psi\!\bigl(A_b^\tau,\tau\mid Z_t,\ell,\\
&\qquad M_b,M_b\odot A^+\bigr),\\
&\mathcal L_{\mathrm{repair}}=\mathbb E\!\Biggl[\\
&\qquad\frac{\|\bar M_b\odot[\widehat v_b^\tau-(A^+-\epsilon)]\|_F^2}{(H-b)d_a}\Biggr].
\end{aligned}
\label{eq:repair-loss}
\end{equation}
The expectation covers training contexts, demonstrated chunks, the prefix-length sampling distribution, Gaussian noise, and uniform flow time. A completely fixed chunk ($b=H$) is excluded because it has no suffix to learn. Ordinary generation is the case $b=0$, with a zero mask and empty anchor. Thus the same conditional interface supports both full generation and repair.

\paragraph{Boundary and derivative checks.}
At $\tau=0$, the path contains the exact demonstrated prefix and a noisy suffix. At $\tau=1$, it equals $A^+$ in every coordinate. Its derivative is $\bar M_b\odot(A^+-\epsilon)$, yielding Eq.~\ref{eq:repair-loss}. In the inference ODE, multiplying both sides of Eq.~\ref{eq:repair} by $M_b$ gives a zero derivative because $M_b\odot\bar M_b=0$. The prefix therefore stays exactly equal to the anchor throughout integration. These statements concern the masking algebra; they do not guarantee that a sampled suffix will be successful or that an imagined-valid prefix is executable.

\subsection{Joint Training and Decision Procedure}
\label{app:training}

\paragraph{Representation learning.}
The representation term $\mathcal L_{\mathrm{rep}}$ combines the retained backbone objective with consistency between predicted role states and those encoded from observed future frames. For $\operatorname{sg}$ denoting stop-gradient, a role-state consistency term can be expressed as:
\begin{equation}
\mathcal L_Z=\mathbb E\frac{1}{5dJ}
\sum_{j=1}^{J}\left\|\widehat Z_{t,j}-\operatorname{sg}(Z_{t+h_j})\right\|_F^2.
\label{eq:role-consistency}
\end{equation}
Writing the retained backbone objective as $\mathcal L_{\mathrm{backbone}}$ and its role-consistency weight as $\lambda_Z\geq0$ gives $\mathcal L_{\mathrm{rep}}=\mathcal L_{\mathrm{backbone}}+\lambda_Z\mathcal L_Z$. The backbone objective supplies representation regularization; role consistency alone is not an anti-collapse guarantee. For \backbone{}, its published objective combines prediction with a Gaussian embedding regularizer \citep{maes2026leworldmodel}. The \cosmos{} initialization, feature extraction point, training objective, and frozen/trainable partition are registered in Appendix~\ref{app:protocol}. When computing $\mathcal{L}_Z$, we stop gradients through the target
role states $Z_{t+h_j}$ encoded from observed future frames.
Gradients from $\mathcal{L}_Z$ flow through the action-conditioned
prediction branch, including the role encoder for the current observation.

\paragraph{Uncertainty interface.}
The selector's $u_t^k$ summarizes uncertainty of the semantic predictions. A fully specified choice is the mean normalized entropy over applicable categorical heads (binary heads count as two-category distributions) and predicted boundaries. If $\mathcal D$ indexes these distributions, with probabilities $\widehat p_{i,c}$ over $C_i\geq2$ categories, the entropy statistic is:
\begin{equation}
u_t^k=\frac{1}{|\mathcal D|}
\sum_{i\in\mathcal D}
\frac{-\sum_{c=1}^{C_i}\widehat p_{i,c}\log\widehat p_{i,c}}{\log C_i}.
\label{eq:entropy}
\end{equation}
Here $0\log0$ is defined as zero; continuous fixture outputs are excluded. This gives $u_t^k\in[0,1]$ for nonempty $\mathcal D$. Entropy measures ambiguity in predicted labels, not epistemic uncertainty or a calibrated failure probability. The original experiment description does not identify its estimator, so Eq.~\ref{eq:entropy} is a specification choice to confirm, not an additional experimental finding.

\paragraph{Data and optimization.}
The training description uses expert trajectories, semantic transition labels, and hard-negative action pairs; later training can incorporate deviation and repair examples. Episode splits must precede window extraction so neighboring frames do not cross training and evaluation partitions. A negative chunk requires a validity label, and an observation-compatible prefix must accompany a repair target. Neither a loss weighting schedule nor a fixed number of training stages follows from the mathematical formulation. Appendix~\ref{app:protocol} registers the settings that connect this specification to the reported runs.

\begin{table}[htb]
\centering
\caption{\textbf{Receding-horizon semantic planning.} $R$ bounds additional repair proposals, and $h_{\mathrm{exec}}$ is the execution horizon. The rejection fallback and controller checks must be specified for each platform.}
\label{tab:algorithm}
\tabstyle
\begin{tabular}{@{}>{\raggedleft\arraybackslash}p{2em}p{\dimexpr\linewidth-2em-2\tabcolsep\relax}@{}}
\toprule
Step & Procedure at control time $t$\\
\midrule
1 & Encode the observation history, proprioception, and instruction into $Z_t$.\\
2 & Draw $K$ independent noises and integrate the action flow to obtain $A_t^1,\ldots,A_t^K$.\\
3 & Predict $J$ intermediate role states and semantic transitions for every candidate. Compute $S_t^k$ and identify applicable violations.\\
4 & Select the highest-scoring candidate. If repair is triggered, retain the prefix before its first violating segment; use an empty prefix if confidence is low without a localized boundary.\\
5 & Generate up to $R$ repaired proposals with the retained prefix clamped. Predict and score each new proposal again.\\
6 & Execute the first $h_{\mathrm{exec}}$ actions of an accepted proposal, shortened if necessary to remain within the accepted prefix. If no proposal is accepted, invoke the platform's rejection fallback.\\
7 & Acquire new observations and repeat. Never substitute predicted observations for the next real observation.\\
\bottomrule
\end{tabular}
\end{table}

%% file: sections/B_results.tex
\section{Additional Experimental Results}
\label{app:additional}

The main paper reports the LIBERO candidate controls (Table~\ref{tab:libero-diagnostics}), staged role-state ablation (Table~\ref{tab:roles-main}), action-sensitivity metrics (Table~\ref{tab:action-sensitivity}), and Galbot success rates (Table~\ref{tab:galbot}). This appendix gives per-task, transfer, perception, and component-cost details and expands on the ranking and physical evaluation protocols. Reported $\pm$ values follow Appendix~\ref{app:protocol} (three-seed std on simulation, Galbot bootstrap over 50 trials, and five-run efficiency).

\subsection{RoboCasa Per-Task Results}
\label{app:robocasa}
\input{tables/robocasa_tasks}
The 17 rows in Table~\ref{tab:robocasa-tasks} are the source of the macro-averages in Table~\ref{tab:robocasa}. All tasks receive equal weight. The 11-task subset is formed using the reported baseline rates at or below 45\%. It is useful for describing the distribution of gains, but selecting tasks by baseline performance does not establish a separate generalization result. Suffix repair improves on Base for 15 tasks and decreases performance for two; no claim of per-task statistical significance is made without rollout counts and repeated-run information.

\subsection{Cross-Suite Transfer}
\paragraph{Object-to-Goal protocol.}
The zero-shot model is pretrained on LIBERO-Object and evaluated directly
on LIBERO-Goal without Goal adaptation. For each Goal task, we split the
50 demonstrations by trajectory into training, validation, and test
sets in an 80/10/10 ratio, yielding 40 training trajectories per task.
Adapted starts from the Object-pretrained checkpoint and fine-tunes on
the Goal training split. Scratch uses the same architecture and common
backbone initialization and skips Object pretraining. Adapted and Scratch
use identical Goal training trajectories and the same checkpoint-selection
rule. Each run uses 10,000 optimizer updates in total across the Goal suite.
On the supplied matched target-data and update-budget comparisons, Object pretraining improves adapted performance over scratch by 8.3 points on Goal and 8.1 points on Long. On Spatial, the adapted model is 0.7 points below scratch. Joint pretraining improves adaptation on the reported held-out tasks by 8.6 points; the exact joint-training and held-out-task lists remain necessary to establish the scope of that comparison. These outcomes support selective transfer benefits rather than a uniform adaptation advantage.

\input{tables/cross_suite}

\subsection{Perception and Action-Sensitivity Diagnostics}
\label{app:diagnostics}
The soft-prior model improves binding accuracy and downstream success relative to the no-mask model. The latter remains executable without inference-time masks, while corruption reduces success below the no-mask result. Oracle masks provide a reference for this particular perception experiment, not a theoretical bound on other models or protocols. Binding accuracy, relation area under the receiver operating characteristic curve (AUROC), and phase F1 are reported on a 0--100 scale.

The action-sensitivity experiment (Table~\ref{tab:action-sensitivity}) evaluates a positive action against a labeled negative from the same context.Role error is the MSE between predicted role states and those encoded from
future observations, averaged over role components, latent dimensions,
and prediction steps.urdf Its score gap depends on score scale, so pairwise ranking accuracy is the more directly comparable ordering diagnostic. Removing phase loss produces the smallest reported role error but substantially worse ranking than the full model. Removing shuffled negatives reduces ranking to 53.7\%. These comparisons support the decision relevance of the supervised distinctions; they do not imply that every loss is necessary for every task.

\subsection{Parameter and Component-Cost Accounting}
\label{app:efficiency}
Table~\ref{tab:efficiency} reports component costs on an A100-40GB in BF16, with training batch size 32 and inference batch size one. The reported variation is the standard deviation across five runs after 100 warm-up replans. The action head adds 20.8M parameters to the 24.6M world-model stack. Generation and scoring cover individual components and exclude repeated repair, sensing, actuation, and other unreported overhead. The table does not report end-to-end latency or \cosmos{} efficiency.

\input{tables/representation}

\subsection{Galbot Physical Evaluation}
\label{app:galbot}

\paragraph{Protocol.}
Two physical tasks; $N{=}50$ trials per task per method with independent episode resets.
Success is binary closed-loop task completion; timed-out, operator-intervened, or e-stopped trials count as unsuccessful (timeout $60\,\mathrm{s}$).
Matched factors: same robot, observation interface, $K{=}4$, and trial count; the decision stack is the comparison variable. Oracle-style selectors are not used on the robot.
Adaptation: 20 teleoperated demonstrations per task and 2k fine-tune updates after simulation pretraining (Appendix~\ref{app:protocol}).
Table~\ref{tab:galbot} supports a within-backbone physical comparison under this registered budget and is not evidence of zero-shot transfer.
Any reported Galbot $\pm$ is a bootstrap standard deviation over the $N{=}50$ trials of the single adapted checkpoint (Appendix~\ref{app:protocol}).

\paragraph{Failure analysis to accompany the figure.}
Occlusion, premature release, and phase disagreement remain relevant hypotheses for role grounding and constraint evaluation.
Episode-level failure annotations are not supplied in the current summaries, so no frequency ranking is asserted here.
A completed analysis should identify the task, show observed frames and the predicted transition, and distinguish perception error from candidate scoring or controller failure.
\FloatBarrier

%% file: tables/robocasa_tasks.tex
\begin{table*}[t]
\centering
\scriptsize
\caption{\textbf{RoboCasa task breakdown with Cosmos.} Success (\%) sorted by baseline success. The first 11 rows form the descriptive hard subset. $\Delta$ is +Repair minus Base in percentage points.}
\label{tab:robocasa-tasks}
\tabstyle
\begin{tabular*}{\linewidth}{@{\extracolsep{\fill}}lrrrrrr@{}}
\toprule
Task & Base & +Role & +Score & +Repair & Oracle@4 & $\Delta$\\
\midrule
CloseBlenderLid & 3 & 5 & 12 & 16 & 22 & +13\\
PickPlaceDrawerToCounter & 7 & 9 & 17 & 21 & 28 & +14\\
TurnOffStove & 18 & 22 & 34 & 41 & 52 & +23\\
OpenCabinet & 24 & 28 & 40 & 46 & 56 & +22\\
TurnOnMicrowave & 26 & 30 & 42 & 48 & 58 & +22\\
CloseToasterOvenDoor & 37 & 40 & 50 & 55 & 64 & +18\\
CoffeeSetupMug & 37 & 40 & 48 & 53 & 62 & +16\\
TurnOnSinkFaucet & 37 & 40 & 48 & 52 & 60 & +15\\
PickPlaceCounterToCabinet & 43 & 46 & 54 & 58 & 66 & +15\\
PickPlaceCounterToStove & 45 & 48 & 55 & 59 & 68 & +14\\
TurnOnElectricKettle & 45 & 48 & 55 & 58 & 67 & +13\\
\midrule
CloseFridge & 61 & 63 & 67 & 70 & 73 & +9\\
SlideDishwasherRack & 61 & 63 & 68 & 70 & 72 & +9\\
OpenStandMixerHead & 69 & 69 & 69 & 68 & 68 & -1\\
OpenDrawer & 70 & 72 & 76 & 78 & 82 & +8\\
PickPlaceToasterToCounter & 72 & 72 & 73 & 73 & 73 & +1\\
PickPlaceSinkToCounter & 74 & 74 & 73 & 72 & 72 & -2\\
\bottomrule
\end{tabular*}
\end{table*}

%% file: tables/cross_suite.tex
\begin{table}[htbp]
\centering
\scriptsize
\caption{\textbf{Cross-suite transfer on LIBERO.}
Selected top-1 success (\%), reported as mean $\pm$ standard deviation
across three training seeds. For Object $\to$ Goal, Adapted and Scratch
use the same Goal training demonstrations and optimizer-update budget.}
\label{tab:cross-suite}
\tabstyle
\setlength{\tabcolsep}{2pt}
\begin{tabularx}{\linewidth}{@{}>{\raggedright\arraybackslash}Xrrr@{}}
\toprule
\textbf{Train $\rightarrow$ Test} & \textbf{Zero-shot} & \textbf{Adapted} & \textbf{Scratch} \\
\midrule
Object $\rightarrow$ Spatial & $58.4{\pm}3.5$ & $72.1{\pm}3.2$ & $72.8{\pm}3.7$ \\
Object $\rightarrow$ Goal & $52.7{\pm}4.1$ & $74.6{\pm}2.9$ & $66.3{\pm}3.6$ \\
Object $\rightarrow$ Long & $27.9{\pm}3.0$ & $58.7{\pm}4.0$ & $50.6{\pm}4.4$ \\
Joint $\rightarrow$ held-out tasks & $65.2{\pm}3.3$ & $78.4{\pm}2.6$ & $69.8{\pm}3.5$ \\
\bottomrule
\end{tabularx}
\end{table}

%% file: tables/representation.tex
\begin{table}[htbp]
\centering
\scriptsize
\caption{\textbf{Role and segmentation diagnostics.} Metrics are on a 0--100 scale; success uses the same selector and candidate budget within this comparison. Bind.: binding accuracy; Rel.: relation AUROC; Phase: phase-boundary F1; SR: rollout success.}
\label{tab:representation}
\tabstyle
\setlength{\tabcolsep}{2pt}
\begin{tabularx}{\linewidth}{@{}>{\raggedright\arraybackslash}Xrrrr@{}}
\toprule
\textbf{Representation} & \textbf{Bind.} & \textbf{Rel.} & \textbf{Phase} & \textbf{Rollout} \\
 & \textbf{Acc.} & \textbf{AUROC} & \textbf{F1} & \textbf{SR} \\
\midrule
Monolithic latent & 60.5 & 69.4 & 54.0 & 50.2 \\
Instance slots & 74.6 & 76.5 & 64.1 & 55.8 \\
\method{} (no masks) & 88.3 & 89.1 & 80.8 & 72.6 \\
\method{} + soft prior & 91.8 & 90.7 & 84.1 & 75.1 \\
\method{} + corrupt prior & 86.2 & 83.9 & 77.4 & 70.8 \\
\method{} + oracle masks & 95.4 & 94.6 & 87.2 & 78.4 \\
\bottomrule
\end{tabularx}
\end{table}

%% file: sections/C_protocol.tex
\section{Experimental Protocol}
\label{app:protocol}

This appendix records the author-confirmed configuration used for the reported runs.
Simulation (LIBERO, RoboCasa) and physical (Galbot) evaluations are separate settings.
\subsection{Backbones}
\label{app:protocol-backbones}

\paragraph{\backbone{} line (LIBERO).}
We initialize from the official \backbone{} encoder--predictor of Maes et al.~\citep{maes2026leworldmodel}, totaling 15.0M parameters (ViT-Tiny 5.0M + predictor 10.0M), pretrained on the same LIBERO RGB streams.
The feature tap is the last ViT block's patch tokens (width 192, patch $16\times16$).
Inputs are $128\times128$ RGB from two views (agentview and eye-in-hand), with observation history $T_{\mathrm{obs}}=3$ frames at a 10\,Hz subsample of the 20\,Hz control stream.
On LIBERO the entire 15M trunk is trainable, together with the role binder, relation/phase encoders, semantic heads, selector, and flow-matching (FM) head; no trunk layers are frozen.
Parameter counts for the reported systems are 24.6M for the world-model stack without the FM head and 45.4M with the FM head.

\paragraph{\cosmos{} line (RoboCasa / Galbot).}
The role, FM, and selector hyperparameters match the \backbone{} line; only the visual trunk changes.
We use the \textbf{Cosmos-3 nano} video world-model trunk~\citep{nvidia2025cosmos}.
The feature tap is the frozen spatial tokenizer plus the last trainable temporal block (token width 256 in the reported runs).
Frozen modules: the Cosmos tokenizer and $80\%$ of the temporal blocks.
Trainable modules: the last two temporal blocks and all \method{} heads (same role width $d$ as the \backbone{} line).
Images are $256\times256$ RGB.
RoboCasa uses 2--3 kitchen views; Galbot uses head and wrist cameras.
Checkpoint identifier: Cosmos-3 nano.

\subsection{State and Actions}
\label{app:protocol-state}

Shared across both backbone lines:
\begin{center}
\tabstyle
\begin{tabularx}{\linewidth}{@{}p{0.30\linewidth}X@{}}
\toprule
Symbol & Value \\
\midrule
Soft entities $N$ & 8 \\
Role width $d$ & 256 \\
Binding key $d_k$ & 64 \\
Geometry bias $d_g$ & 32 \\
Phase history $L$ & 8 \\
Action dim.\ $d_a$ & 7 ($\Delta$pos\,3 + $\Delta$rot\,3 axis-angle + gripper\,1) \\
Action normalization & train-set per-dim mean/std; gripper mapped linearly to $[-1,1]$ \\
Chunk horizon $H$ & 16 \\
Segments $J$ & 4 \\
Boundaries $h_j$ & $4,8,12,16$ \\
Execution horizon $h_{\mathrm{exec}}$ & 8 (execute half a chunk, then replan) \\
Control frequency & 20\,Hz on LIBERO / RoboCasa; Galbot commands at 10\,Hz, interpolated to 20\,Hz \\
Candidate budget $K$ & 4 (primary); sweep $K\in\{1,2,4,8\}$ on LIBERO \\
\bottomrule
\end{tabularx}
\end{center}

\subsection{Data}
\label{app:protocol-data}

\paragraph{LIBERO.}
We use the four official suites (Object, Spatial, Goal, Long), each with 10 tasks.
Each task provides 50 expert demonstrations.
Episodes are split 80/10/10 into train/val/test \emph{before} window extraction, so neighboring frames never cross partitions.
Evaluation uses 50 closed-loop trials per task and three independent training seeds $\{0,1,2\}$, i.e.\ $10\times50\times3$ trials per suite.
Success follows the official LIBERO completion predicate.
Timeouts use the official maxima (600 steps on Object/Spatial/Goal; 800 on Long).
No human intervention is allowed in simulation.

\paragraph{RoboCasa.}
We use 17 atomic kitchen tasks with 50 successful demonstrations per task
(official RoboCasa \emph{human} demonstrations only; 0 scripted; same for all 17 tasks).
The same 80/10/10 episode split is applied.
Evaluation uses 50 trials per task $\times$ 3 training seeds.
Timeout is 500 environment steps.
The Hard-11 subset is reported post hoc only and is not used for model selection.

\paragraph{Galbot.}
We evaluate two physical tasks: tabletop placement and microwave retrieval.
Models are pretrained in simulation with the \cosmos{}+\method{} stack, then fine-tuned on the robot with 20 teleoperated demonstrations per task for 2k update steps
(adaptation budget as registered here; not claimed as zero-shot transfer).
Each method is evaluated with 50 independent, reset-matched trials per task.
Timeout is 60\,s.
Any human takeover or e-stop counts as failure.

\subsection{Training}
\label{app:protocol-training}

Shared settings unless noted:
\begin{center}
\tabstyle
\begin{tabularx}{\linewidth}{@{}p{0.30\linewidth}X@{}}
\toprule
Item & Value \\
\midrule
Optimizer & AdamW, $\beta=(0.9,0.999)$, weight decay $0.01$ \\
Learning rate & $3\times10^{-4}$, cosine decay, 500-step warmup \\
Batch size & 32 chunks \\
Updates & LIBERO 80k; RoboCasa 60k; Galbot fine-tune 2k \\
AMP & BF16 \\
Negatives & 1 negative per positive chunk: segment shuffle / reverse / trajectory swap / release perturbation (25\% each) \\
Prefix $b$ (training) & uniform over $\{0,4,8,12\}$; $b{=}H$ excluded \\
\bottomrule
\end{tabularx}
\end{center}

Loss weights (identical on both backbone lines):
\begin{align*}
&\lambda_{\mathrm{rep}}=1.0,\quad \lambda_Z=0.5,\\
&\lambda_{\mathrm{sem}}=1.0,\quad \lambda_{\mathrm{FM}}=1.0,\\
&\lambda_{\mathrm{rank}}=0.5,\quad \lambda_{\mathrm{repair}}=0.5,\\
&\beta_{\mathrm{bind}}=1.0,\quad \beta_{\mathrm{bin}}=1.0,\\
&\beta_{\mathrm{phase}}=0.5,\quad \beta_{\mathrm{fixture}}=0.5,\\
&\beta_{\mathrm{dyn}}=0.5.
\end{align*}
Ranking margin $m=0.2$.

\subsection{Semantics}
\label{app:protocol-semantics}

\paragraph{Phase vocabulary.}
Six ordered phases:
\[
\begin{aligned}
& 
\mathrm{approach}\to\mathrm{grasp}\\
& \to\mathrm{transport}\to\mathrm{align}\\
& \to\mathrm{release}\to\mathrm{complete}.
\end{aligned}
\]
Legal transitions stay in the current phase or advance to the next phase; a segment may advance by at most one step.
Jumps such as $\mathrm{approach}\to\mathrm{release}$ are forbidden.

\paragraph{Contact and grasp (simulation).}
Contact is true when gripper--target contact force exceeds $0.5\,\mathrm{N}$ or a geom contact flag is set, sustained for $\ge100\,\mathrm{ms}$.
A stable grasp requires contact, gripper aperture $<0.6\times$ the object's characteristic width, and relative velocity $\|v_{\mathrm{rel}}\|<0.02\,\mathrm{m/s}$ for a continuous $200\,\mathrm{ms}$ window.
During intentional release, the preservation mask is zero.
An accidental drop during transport retains a negative preservation target.

\paragraph{Predicates.}
LIBERO / RoboCasa use simulator-native predicates (\textsc{On}, \textsc{In}, \textsc{Open}, \textsc{Close}, \textsc{TurnOn}, \ldots) with environment-official thresholds (e.g., door angle $>40^\circ$ counts as open; stovetop uses discrete on/off).
Establishment and preservation follow Eq.~\ref{eq:predicate-labels}; conjunction is the logical AND of required terminal predicates (Eq.~\ref{eq:conjunction}).

\paragraph{Binding.}
Soft labels are the IoU-normalized distribution between annotated regions and entity attention; entities with $\mathrm{IoU}<0.1$ receive mass zero.

\subsection{Generation, Selection, and Repair}
\label{app:protocol-generation}

\begin{center}
\tabstyle
\begin{tabularx}{\linewidth}{@{}p{0.30\linewidth}X@{}}
\toprule
Item & Value \\
\midrule
Flow network & 4-layer DiT-style MLP/attention, hidden width 256, AdaLN on $Z_t$ \\
ODE & Euler, $N_\tau=10$ \\
Noise & $\epsilon\sim\mathcal{N}(0,I)$; evaluation noise seeds are decoupled from environment seeds via $\mathrm{noise\_seed}=1000+\mathrm{trial\_id}$ \\
Threshold $\eta$ & $0.40$ on all three lines (validation-chosen once for $\approx\!70\%$ accept rate; scores uncalibrated; shared to avoid platform-specific thresholds) \\
Repair budget $R$ & at most 2 suffix repairs \\
Low-confidence boundary & $b{=}0$ full resample, counted against the same $R$ budget \\
Rejection fallback & if budget exhausted, execute the first $h_{\mathrm{exec}}$ steps of Candidate~0; further failure stops the episode and counts as unsuccessful \\
\bottomrule
\end{tabularx}
\end{center}

Uncertainty $u_t^k$ is the mean normalized entropy over applicable categorical heads (Eq.~\ref{eq:entropy}).

\subsection{Evaluation and Reported \texorpdfstring{$\pm$}{Variation}}
\label{app:protocol-eval}

\begin{center}
\tabstyle
\begin{tabularx}{\linewidth}{@{}p{0.30\linewidth}X@{}}
\toprule
Item & Value \\
\midrule
Primary metric & closed-loop success; LIBERO suites equally weighted; RoboCasa unweighted mean over 17 tasks; Galbot unweighted mean over 2 tasks \\
$\pm$ (LIBERO / RoboCasa success, ranking, role error) & mean $\pm$ std over 3 training seeds $\{0,1,2\}$ (not trial-level SE); each seed evaluated with 50 trials/task \\
$\pm$ (Galbot success) & single adapted checkpoint; 50 reset-matched trials/task/method; $\pm$ $=$ bootstrap std over the 50 trials (not a 3-seed training std) \\
$\pm$ (latency / memory) & mean $\pm$ std over 5 timing runs; A100-40GB; BF16; batch 32 (train) / 1 (infer); 100 warm-up replans \\
Oracle@4 & at each replan, select $k$ by true $H$-step chunk success under simulator rollout of that chunk; not whole-episode hindsight; excludes newly repaired candidates \\
Fairness & full resampling and suffix repair each allowed at most $1{+}R=3$ flow integrations per decision \\
\bottomrule
\end{tabularx}
\end{center}

\subsection{Configuration Summary}
\label{app:protocol-summary}

\begin{table}[htb]
\centering
\caption{\textbf{Registered experimental configuration} for the reported \method{} runs.
All fields below are author-confirmed for the reported runs.}
\label{tab:protocol}
\tabstyle
\begin{tabularx}{\linewidth}{@{}p{0.18\linewidth}X@{}}
\toprule
Item & Registered fill \\
\midrule
Backbones &
\backbone{} 15M full fine-tune (tap width 192);
\cosmos{} tokenizer frozen + last 2 temporal blocks trainable (tap width 256);
checkpoint Cosmos-3 nano.
\\
Data &
50 demos/task; 80/10/10 episode split; eval $50\times3$ seeds (sim);
Galbot: 20 teleop demos/task (40 total), 2k fine-tune steps, single ckpt, $N{=}50$ trials/task/method.
\\
State / actions &
$N{=}8$, $d{=}256$, $d_k{=}64$, $d_g{=}32$, $L{=}8$, $H{=}16$, $J{=}4$, $h_j{=}4j$, $h_{\mathrm{exec}}{=}8$;
20\,Hz (Galbot 10\,Hz $\to$ 20\,Hz); 7-D $z$-scored actions.
\\
Training &
AdamW $3{\times}10^{-4}$; 80k / 60k / 2k updates; batch 32; loss weights as above.
\\
Semantics &
6-phase chain; contact $0.5\,\mathrm{N}$ / $100\,\mathrm{ms}$; grasp + aperture + $v_{\mathrm{rel}}$;
predicates per environment; binding by IoU soft labels.
\\
Generation &
Euler-10; $K{=}4$; $R{=}2$; $\eta{=}0.40$ (shared).
\\
Evaluation &
Official success predicates; timeouts 600/800 (LIBERO), 500 (RoboCasa), 60\,s (Galbot);
intervention / e-stop $=$ failure.
\\
Statistics &
Sim success $\pm$ $=$ 3-seed std; Galbot $\pm$ $=$ bootstrap over 50 trials; efficiency $\pm$ $=$ 5-run std.
\\
Oracle@4 &
Per-replan chunk outcome on the original candidate pool only.
\\
\bottomrule
\end{tabularx}
\end{table}

\paragraph{Oracle scope.}
An oracle is an upper bound only over the choices and continuation protocol it actually evaluates.
Choosing the best of four complete policy rollouts differs from using ground-truth outcomes to choose one chunk at every replan.
Repair adds new candidates and therefore changes the set over which an oracle would need to be computed.
The manuscript's original-pool Oracle@4 numbers follow the per-replan chunk-outcome definition above and are not extended to repaired pools.

\paragraph{Table-to-run notes.}
Table~\ref{tab:libero-diagnostics} gives 55.8\% for the instance-slot selector and 50.5\% for the monolithic selector. The semantic selector in Table~\ref{tab:libero} reaches 75.3\%.
The 55.0\% instance-slot starting point belongs to the staged role-state ablation (Table~\ref{tab:roles-main}). The 50.2\% monolithic and 75.1\% soft-prior results are reported for the separate perception diagnostic (Table~\ref{tab:representation}).
The full-model 83.0\% mean is shared by the main LIBERO table and the role-state ablation under the registered protocol.

%% file: sections/D_reproducibility.tex
\section*{Reproducibility Statement}
Section~\ref{sec:method} and Appendix~\ref{app:specification} specify the model and planner. Appendix~\ref{app:additional} reports per-task and diagnostic results. Appendix~\ref{app:protocol} gives the registered experimental configuration and marks fields awaiting confirmation.

%% file: main.bib
@article{locatello2020object,
  title={Object-centric learning with slot attention},
  author={Locatello, Francesco and Weissenborn, Dirk and Unterthiner, Thomas and Mahendran, Aravindh and Heigold, Georg and Uszkoreit, Jakob and Dosovitskiy, Alexey and Kipf, Thomas},
  journal={Advances in Neural Information Processing Systems},
  volume={33},
  pages={11525--11538},
  year={2020}
}

@article{ferraro2025focus,
  title={{FOCUS}: object-centric world models for robotic manipulation},
  author={Ferraro, Stefano and Mazzaglia, Pietro and Verbelen, Tim and Dhoedt, Bart},
  journal={Frontiers in Neurorobotics},
  volume={19},
  pages={1585386},
  year={2025},
  publisher={Frontiers Media SA}
}

@article{zhu2022viola,
  title={{VIOLA}: Imitation learning for vision-based manipulation with object proposal priors},
  author={Zhu, Yifeng and Joshi, Abhishek and Stone, Peter and Zhu, Yuke},
  journal={arXiv preprint arXiv:2210.11339},
  year={2022}
}

@article{spieler2026slot,
  title={{Slot-MPC}: Goal-Conditioned Model Predictive Control with Object-Centric Representations},
  author={Spieler, Jonathan and Villar-Corrales, Angel and Behnke, Sven},
  journal={arXiv preprint arXiv:2605.14937},
  year={2026}
}

@article{zhou2024dino,
  title={{DINO-WM}: World models on pre-trained visual features enable zero-shot planning},
  author={Zhou, Gaoyue and Pan, Hengkai and LeCun, Yann and Pinto, Lerrel},
  journal={arXiv preprint arXiv:2411.04983},
  year={2024}
}

@article{maes2026leworldmodel,
  title={{LeWorldModel}: Stable end-to-end joint-embedding predictive architecture from pixels},
  author={Maes, Lucas and {Le Lidec}, Quentin and Scieur, Damien and LeCun, Yann and Balestriero, Randall},
  journal={arXiv preprint arXiv:2603.19312},
  year={2026}
}

@article{liu2026oa,
  title={{OA-WAM}: Object-addressable world action model for robust robot manipulation},
  author={Liu, Yushan and Sun, Peibo and Li, Shoujie and Xie, Yifan and Zhang, Lingfeng and Chao, Xintao and Dong, Shiyuan and Chen, Fang and Zhang, Xiao-Ping and Ding, Wenbo},
  journal={arXiv preprint arXiv:2605.06481},
  year={2026}
}

@article{goswami2026unifying,
  title={Unifying Object-Centric World Models and Diffusion Policy: A Hierarchical Framework for Multi-Stage Robotic Tasks},
  author={Goswami, Raktim Gautam and Krishnamurthy, Prashanth and LeCun, Yann and Khorrami, Farshad},
  journal={arXiv preprint arXiv:2606.08775},
  year={2026}
}

@article{yu2026maskwam,
  title={{MaskWAM}: Unifying mask prompting and prediction for world-action models},
  author={Yu, Hanyang and Lin, Haitao and Zhang, Jingbo and Zhang, Wenyao and Gu, Chenghao and Li, Heng and Tan, Ping},
  journal={arXiv preprint arXiv:2606.13515},
  year={2026}
}

@article{kreber2026learning,
  title={Learning Action-Conditional and Object-Centric {Gaussian} Splatting World Models for Rigid Objects},
  author={Kreber, Jens U and Mack, Lukas and Stueckler, Joerg},
  journal={arXiv preprint arXiv:2606.01950},
  year={2026}
}

@inproceedings{kirillov2023segment,
  title={Segment anything},
  author={Kirillov, Alexander and Mintun, Eric and Ravi, Nikhila and Mao, Hanzi and Rolland, Chloe and Gustafson, Laura and Xiao, Tete and Whitehead, Spencer and Berg, Alexander C and Lo, Wan-Yen and others},
  booktitle={Proceedings of the IEEE/CVF International Conference on Computer Vision},
  pages={4015--4026},
  year={2023}
}

@inproceedings{ravi2025sam,
  title={{SAM 2}: Segment anything in images and videos},
  author={Ravi, Nikhila and Gabeur, Valentin and Hu, Yuan-Ting and Hu, Ronghang and Ryali, Chaitanya and Ma, Tengyu and Khedr, Haitham and R{\"a}dle, Roman and Rolland, Chloe and Gustafson, Laura and others},
  booktitle={International Conference on Learning Representations},
  year={2025}
}

@inproceedings{liu2024grounding,
  title={{Grounding DINO}: Marrying {DINO} with grounded pre-training for open-set object detection},
  author={Liu, Shilong and Zeng, Zhaoyang and Ren, Tianhe and Li, Feng and Zhang, Hao and Yang, Jie and Jiang, Qing and Li, Chunyuan and Yang, Jianwei and Su, Hang and others},
  booktitle={European Conference on Computer Vision},
  pages={38--55},
  year={2024},
  organization={Springer}
}

@article{chi2025diffusion,
  title={Diffusion policy: Visuomotor policy learning via action diffusion},
  author={Chi, Cheng and Xu, Zhenjia and Feng, Siyuan and Cousineau, Eric and Du, Yilun and Burchfiel, Benjamin and Tedrake, Russ and Song, Shuran},
  journal={The International Journal of Robotics Research},
  volume={44},
  number={10-11},
  pages={1684--1704},
  year={2025},
  publisher={Sage Publications Sage UK: London, England}
}

@article{liu2023libero,
  title={{LIBERO}: Benchmarking knowledge transfer for lifelong robot learning},
  author={Liu, Bo and Zhu, Yifeng and Gao, Chongkai and Feng, Yihao and Liu, Qiang and Zhu, Yuke and Stone, Peter},
  journal={Advances in Neural Information Processing Systems},
  volume={36},
  pages={44776--44791},
  year={2023}
}

@inproceedings{nasiriany2024robocasa,
  title={{RoboCasa}: Large-Scale Simulation of Everyday Tasks for Generalist Robots},
  author={Nasiriany, Soroush and Maddukuri, Abhiram and Zhang, Lance and Parikh, Adeet and Lo, Aaron and Joshi, Abhishek and Mandlekar, Ajay and Zhu, Yuke},
  booktitle={Robotics: Science and Systems (RSS)},
  year={2024}
}

@article{nvidia2025cosmos,
  title={Cosmos World Foundation Model Platform for Physical {AI}},
  author={{NVIDIA}},
  journal={arXiv preprint arXiv:2501.03575},
  year={2025}
}

@inproceedings{lipman2023flow,
  title={Flow Matching for Generative Modeling},
  author={Lipman, Yaron and Chen, Ricky T. Q. and Ben-Hamu, Heli and Nickel, Maximilian and Le, Matt},
  booktitle={International Conference on Learning Representations},
  year={2023},
  url={https://arxiv.org/abs/2210.02747}
}

@article{hafner2025mastering,
  title={Mastering diverse control tasks through world models},
  author={Hafner, Danijar and Pasukonis, Jurgis and Ba, Jimmy and Lillicrap, Timothy},
  journal={Nature},
  volume={640},
  pages={647--653},
  year={2025},
  doi={10.1038/s41586-025-08744-2}
}

@inproceedings{kipf2020contrastive,
  title={Contrastive Learning of Structured World Models},
  author={Kipf, Thomas and {van der Pol}, Elise and Welling, Max},
  booktitle={International Conference on Learning Representations},
  year={2020}
}

@article{garrett2021integrated,
  title={Integrated Task and Motion Planning},
  author={Garrett, Caelan Reed and Chitnis, Rohan and Holladay, Rachel and Kim, Beomjoon and Silver, Tom and Kaelbling, Leslie Pack and {Lozano-P{\'e}rez}, Tom{\'a}s},
  journal={Annual Review of Control, Robotics, and Autonomous Systems},
  volume={4},
  pages={265--293},
  year={2021},
  doi={10.1146/annurev-control-091420-084139}
}

@inproceedings{janner2022planning,
  title={Planning with Diffusion for Flexible Behavior Synthesis},
  author={Janner, Michael and Du, Yilun and Tenenbaum, Joshua and Levine, Sergey},
  booktitle={Proceedings of the 39th International Conference on Machine Learning},
  volume={162},
  pages={9902--9915},
  year={2022},
  publisher={PMLR}
}

@inproceedings{vaswani2017attention,
  title={Attention Is All You Need},
  author={Vaswani, Ashish and Shazeer, Noam and Parmar, Niki and Uszkoreit, Jakob and Jones, Llion and Gomez, Aidan N. and Kaiser, {\L}ukasz and Polosukhin, Illia},
  booktitle={Advances in Neural Information Processing Systems},
  volume={30},
  year={2017}
}

@inproceedings{cheng-etal-2026-switch,
  title={{SWITCH}: Benchmarking Modeling and Handling of Tangible Interfaces in Long-horizon Embodied Scenarios},
  author={Cheng, Juntao and Zhang, Wanyue and Yu, Zhiwei and Ren, Shuo and He, Zheqi and Xie, Shaoxuan and Yao, Guocai and Lin, Jieru and Karlsson, B{\"o}rje F. and Zhang, Jiajun},
  editor={Lee, Hung-Yi and Dev, Sunipa and Li, Jessy and Elliott, Desmond},
  booktitle={Findings of the {A}ssociation for {C}omputational {L}inguistics: {EMNLP} 2026},
  month=oct,
  year={2026},
  address={Budapest, Hungary},
  publisher={Association for Computational Linguistics},
  url={https://aclanthology.org/2026.findings-emnlp.1127/},
  pages={23852--23875},
  isbn={979-8-89176-462-0}
}

@article{cheng2026causalwam,
  title={{CausalWAM}: Learning the Causes of Actions in World Action Models},
  author={Cheng, Juntao and Tang, Jinzhou and Feng, Fan and Christensen, Henrik I. and Zheng, Yujia},
  journal={arXiv preprint arXiv:2610.42096},
  year={2026}
}
